\documentclass[journal]{IEEEtran}

\usepackage[english]{babel} 
\usepackage[numbers, sort, comma, square]{natbib}
\usepackage{amsmath}
\usepackage{amsfonts}
\usepackage{filecontents}
\usepackage[pdftex]{graphicx}
\usepackage{color}
\usepackage[caption=false,font=footnotesize]{subfig}
\usepackage{booktabs}
\usepackage{multirow}
\usepackage{rotating}
\usepackage{flexisym}
\usepackage{fixltx2e}
\usepackage{url}
\usepackage[pdftex,colorlinks=false,bookmarksopen=true,bookmarksopenlevel=0,pdfborder={0 0 0},breaklinks]{hyperref}
\usepackage[nolist]{acronym}
\usepackage{siunitx}
\usepackage{bm}
\usepackage{ifmtarg}
\usepackage{wasysym}                   
\usepackage{placeins}  

\title{Leveraging over intact priors for boosting control and dexterity of prosthetic hands by amputees}

\author{Valentina~Gregori
        and~Barbara~Caputo
\thanks{V. Gregori is with Department of Computer, Control, and Management Engineering, University of
Rome La Sapienza, via Ariosto 25, 00185 Roma, Italy.}
\thanks{B. Caputo is with Department of Computer, Control, and Management Engineering, University of
Rome La Sapienza, via Ariosto 25, 00185 Roma, Italy.}}

\begin{document}


\maketitle

\begin{abstract}
Non-invasive myoelectric prostheses require a long training time to obtain satisfactory control dexterity. These training times could possibly be reduced by leveraging over training efforts by previous subjects. So-called domain adaptation algorithms formalize this strategy and have indeed been shown to significantly reduce the amount of required training data for intact subjects for myoelectric movements classification. It is not clear, however, whether these results extend also to amputees and, if so, whether prior information from amputees and intact subjects is equally useful.
To overcome this problem, we evaluated several domain adaptation algorithms on data coming from both amputees and intact subjects. 
Our findings indicate that: (1) the use of previous experience from other subjects allows us to reduce the training time by about an order of magnitude; (2) this improvement holds regardless of whether an amputee exploits previous information from other amputees or from intact subjects.\\
\end{abstract}%


\IEEEpeerreviewmaketitle

\section{Introduction}

\IEEEPARstart{A}{} majority of upper-limb amputees is interested in prostheses controlled via surface electromyography (EMG), but they perceive the prohibitively difficult control as a great concern~\cite{atkins96}. This becomes even more problematic with highly articulated modern prostheses. The natural use of these devices is challenging in everyday life primarily due to the software~\cite{zecca02, micera10, peerdeman11}. The open question is how to reduce this training time while making control as natural as possible.
Machine learning has opened a new path to tackle this problem by allowing the prosthesis to adapt to the myoelectric signals of a specific user. Although these methods have been applied with success (e.g., \citep{castellini09} and references therein), they still require a significant amount of data from individual subjects to learn models with satisfactory performance.

Consider a situation in which different subjects repeat the same hand postures and suppose that a new target user attempts to learn the same movements. In this case it should be beneficial to reuse the information from the latter subjects and thereby reduce the training data required from a new subject. However, even if a movement appears the same for all subjects, the distribution of their myoelectric signals is very different. These depend on user characteristics like age, subcutaneous fat, muscular activity and on the position of the electrodes during the data collection~\cite{farina02}. Domain adaptation methods can be used to overcome this mismatch between these different data distributions by considering each subject as an individual domain. This allows boosting prosthetic control by exploiting earlier training efforts by other source subjects.

These techniques have shown promising results for myoelectric movements classification with a large amount of intact subjects~\cite{tommasi11, patricia-tommasi14}. However, these methods have never been verified on a significant number of amputees. Would the findings presented so far still be valid in these cases, or would the differences in the signal caused by the amputation be so strong as to prevent any useful knowledge transfer? And would it be possible to leverage over intact sources to boost control of amputees?

This paper aims at answering these questions. We assess if and how previous knowledge from intact subjects or amputees can reduce the time needed by an amputee to learn to control the device. To this end, we ran three experiments. In the first we considered only intact subjects. Then, we took into account amputated subjects that exploit previous knowledge either from other amputees or from intact subjects (second and third experiment). In order to clarify the importance of prior knowledge, we employed several existing adaptive learning algorithms using EMG data from the largest public database (\url{NinaPro}~\cite{atzoriScData14, atzori12, atzori12tnsre}).

The rest of the paper is organized as follows. In section \ref{sec:RelWork} we summarize the main contributions of previous work. Subsequently we introduce the algorithms used in this work (section \ref{sec:Algorithms}), the data and their processing (section \ref{ExperimentalSetup}). In section \ref{sec:Experiment} we present our results. Section \ref{sec:Conclusion} is a conclusive summary that focuses on the main findings.

\section{Related work} \label{sec:RelWork}
One of the first work that attempted to classify EMG signals of three volunteers is by \citet{graupe1975}. In the following years studies on prosthetic control continued in various directions leading to many advances in the analysis and understanding of EMG. 
Few pioneer works highlight that EMG signals differ from subject to subject and that signals from different subjects are therefore not automatically reusable~\cite{Castellini.etal2009b}. They proposed a cross-subject analysis to show that reusing the pre-trained models from former subjects can shorten the training time for a new user. \citet{matsubara11} proposed to extract a user-independent component from EMG data that is considered transferable across subjects.

Different works applied several domain adaptation methods for the task of EMG recognition, showing that these algorithms outperform non-adaptive ones \cite{sensingero09, orabona09}. \citet{chattopadhyay11} proposed an adaptation method for EMG classification based on a weighted combination of target and source samples. The shortcoming in these studies is that they involved a small number of subjects. Only recent studies~\cite{tommasi11, patricia-tommasi14}, done after the release of the \url{NinaPro} database~\cite{atzoriScData14}, took into account a significant amount of intact subjects (from 10 to 40) and postures (from 12 to 52). They tested several domain adaptation algorithms in various set-ups with different types and numbers of hand movements increasing also the available source subjects. The findings show that posture recognition can be improved by the use of prior knowledge of other intact subjects even with few available training samples. The number of training samples is reduced by an order of magnitude in domain adaptation methods to obtain the same performance achieved as when learning from scratch. Not all methods were found to perform equally well, however. The Two-Stage Weighting Framework~\cite{sun11} method needed a longer running time than the others and the Geodesic Flow Kernel~\cite{gong12} algorithm resulted in poor performance. The methods that achieved the best performance were Multi-Kernel Adaptive Learning~\cite{orabona10} and Multi Adapt~\cite{tommasi11, tommasiOC14}.

\section{Algorithms} \label{sec:Algorithms}
We, first, describe the mathematical background and the a traditional learning algorithm (\ref{sec:Background}), then we present the domain adaptation methods used in the work (\ref{sec:AdaptiveLearning}).

\subsection{Background} \label{sec:Background}
Let us define a training dataset $D = \left\{ \bm{x}_{i}, y_i \right\}_{i=1}^N$ of $N$ input samples $\bm{x}_i \in \mathcal{X}$ and corresponding labels $y_i \in \mathcal{Y}$. In the context of myoelectric classification, the inputs are the EMG signals and the labels are the movements chosen from a set of $G$ possible classes such that the input space $\mathcal{X} \subseteq \mathbb{R}^d$ and output space $\mathcal{Y} = \left\{1, \ldots, G\right\}$. The goal of a classification algorithm is to find a function $h(\bm{x})$ that, for any future input vector $\bm{x}$, can determine the corresponding output $y$. The parameters of this function are determined by solving an optimization problem over the training data.
Between the algorithms that work in this manner, Support Vector Machines (SVMs) are some of the most popular.

We briefly introduce here Least-Square Support Vector Machines (LS-SVM~\cite{suykensLSSVM}) as the starting point of the algorithms described in the next section. It is a variant of SVM and solves a classification problem (in the one-vs-all version) minimizing the squared error with an equality constraint. Given a vector $\bm{x}$ with unknown label, the output hypothesis $h(\bm{x})$ can be written as $\langle\bm{w}, \phi(\bm{x}) \rangle + b $. In the previous equation, $\bm{w}$ and $b$ are the parameters of the separating hyperplane, computed by the following optimization problem:

\begin {equation}
\label{eq:ls-svm}
\begin{aligned}
&\underset{\bm{w},b}{\text{min}} \left\{ \dfrac{1}{2} \Vert \bm{w} \Vert^{2} + \dfrac{C}{2} \sum\limits_{i=1}^N \xi_{i}^{2} \right\} \\
&\text{subject to:}\quad  y_{i} = \langle\bm{w},\phi(\bm{x}_{i}) \rangle+b+\xi_{i}, \quad \forall \textit{i} \in \{ 1,...,N \}\enspace,\\
\end{aligned}\end {equation}
where $C$ is a regularization parameter and $\bm{\xi}$ denotes the prediction errors.
To obtain a better solution we replaced the original training data $\bm{x}$ with $\phi(\bm{x})$, switching from the input space to a higher dimensional feature space. Usually the function $\phi(\cdot)$ is unknown and we work directly with the kernel function $K(\bm{x}\textprime, \bm{x})$, defined as the dot product $\langle \phi(\bm{x}\textprime), \phi(\bm{x}) \rangle$~\cite{suykensLSSVM}. Here we use a Gaussian kernel: 

\begin {equation}
\label{eq:GausKer}
K(\bm{x}\textprime, \bm{x}) = e^{- \gamma \parallel \bm{x}\textprime - \bm{x} \parallel^{2}}\enspace \text{for}\enspace \gamma > 0\enspace.
\end {equation}

\subsection{Adaptive Learning} \label{sec:AdaptiveLearning}
In the following we suppose to have $K$ different sources, where each source is a classification model for the same set of postures. We introduce domain adaptation algorithms that construct a classification model for the same task for a new target using past experience from the sources.

\subsubsection{Multi Adapt (MA)}
This method aims to find for the target a new separating hyperplane $\bm{w}$ close to a linear combination of the pre-trained source models $\hat{\bm{w}}^{k}$~\cite{tommasi11, tommasiOC14}. We solve the following optimization problem: 

\begin{equation}
\label{eq:OptimMultiAdaMulti}
\begin{aligned}
&\underset{\bm{w},b}{\text{min}}\> \left\{ \dfrac{1}{2} \Big\Vert \bm{w}-\sum\limits_{k=1}^K \beta^{k} \hat{\bm{w}}^{k} \Big\Vert^{2}+\dfrac{C}{2} \sum\limits_{i=1}^N \xi_{i}^{2} \right\} \\
&\text{subject to:}\quad y_{i} = \langle\bm{w}, \phi(\bm{x}_{i})\rangle + b + \xi_{i}\enspace.
\end{aligned}
\end{equation}
The vector $\bm{\beta} = [\beta_{1},\beta_{2},...,\beta_{K}]^{T}$ with $\beta_k \geq 0$ represents the contribution of each source in the solution of the target problem and is obtained optimizing a convex upper bound to the
leave-one-out misclassification loss~\cite{tommasiOC14}. 
The same paper proposes a more general case that consists in the use of different weights for different classes of the same source, e.g., $\beta_{k,g}$ is the weight associated to class $g$ of source $k$. This extension is enough intuitive: it can be reasonable to assume that a user learns a movement better from one subject and another movement from a different subject.
In this work we used this latter version referring to this as Multi Adapt (MA).

\subsubsection{Multi Kernel Adaptive Methods (MKAL)}
This  algorithm exploits a combination of source kernels to build the new target model~\cite{orabona10, orabona12}. Let us define: 

\begin{equation}
\label{eq:WBar_PhiBar}
\begin{aligned}
&\bar{\bm{w}} = [\bm{w}^{0},\bm{w}^{1},...,\bm{w}^{K}] \quad \text{and} \\
&\bar{\phi}(\bm{x},y) = [\phi^{0}(\bm{x},y),\phi^{1}(\bm{x},y),...,\phi^{K}(\bm{x},y)]\enspace,
\end{aligned}
\end{equation}
respectively as the concatenated vectors of the hyperplanes and the map functions in the feature space. These are both composed by $(K + 1)$ blocks: one for each source plus the block labelled with $0$, referred to the original training vectors. $\bm{w}^{k} = \left[\bm{w}^{k}_{1},...,\bm{w}^{k}_{G}\right]$ is the hyperplane belonging to source $k$, it is composed of $G$ blocks, one for each class.
The optimization problem becomes:

\begin{equation}
\label{eq:MinProb_MKAL}
\begin{aligned}
&\underset{\bar{\bm{w}}}{\text{min}} \left\{ \dfrac{\lambda}{2} \parallel \bar{\bm{w}} \parallel^{2}_{2,p} + \dfrac{1}{N} \sum\limits_{i=1}^N \xi_{i} \right\}\\
&\text{subject to}\quad \langle \bar{\bm{w}} , (\bar{\phi} (\bm{x}_{i},y_{i}) - \bar{\phi} (\bm{x}_{i},y) ) \rangle \geq 1 - \xi_{i},\, \forall \textit{i} \  y \neq y_{i}\enspace.
\end{aligned}
\end{equation}
The element $p$ sets the level of sparsity in the solution $\bar{\bm{w}}$ and can vary in the range $(1, 2]$. In the final prediction the contribution coming from each source is weighted to reflect the importance of each kernel.

\subsubsection{High Level-Learning2Learn (H-L2L)}
It is an algorithm composed of two layers~\cite{tommasi08, patricia14}. 
In the first layer we build a classification model with an LS-SVM (Equation \ref{eq:ls-svm}) for the target using a part of the original training vectors ($63 \%$ from each class). We use the remaining training vectors ($37 \%$ for each class) to calculate the confidence score for each source model and the target model.
Thus, given a vector $\bm{x}$, for each output class $g$ we have ($K+1$) scores: one for the target $s^{t}(\bm{x},g)$ and $K$ for the models of the sources $s^{s}(\bm{x},g)$.
In the second layer, the vectors used to train the new model are the concatenation of the confidence scores of the target and the sources, obtained from $37 \%$ of the original training vectors. These are exploited to solve a multiclass LS-SVM problem with a Gaussian kernel (Equation \ref{eq:ls-svm}). 

%
%
\section{Experimental Setup} \label{ExperimentalSetup}
The aim of this work is to tackle a situation where a new target user wants to learn to perform the same hand postures that other source subjects are already able to do.
First, we need a large and public database of EMG data to make the result reliable and reproducible. The raw data are then processed to extract the most relevant characteristics.
Starting from these data and the source knowledge we use domain adaptation methods to build classification models for the target. The quality of each model is estimated in terms of recognition rate.
In the following we describe each step in detail.

\subsection{Data} \label{sec:Data}
Data used in our work were collected in the NinaPro (Non Invasive Adaptive Prosthetics) database~\cite{atzoriScData14} (\url{http://ninapro.hevs.ch/}). According to our knowledge, it is the largest existing public database: 40 intact subjects and 11 amputees with 50 postures.
Twelve \textit{Delsys} electrodes acquired EMG data from the arm with a sampling rate of 2 $kHz$.
Each subject repeated several movements 6 times, each repetition is alternated with the rest posture. To perform the exercise intact subjects used the dominant hand, while amputees mimicked with the missing arm.
The postures that we take into account are 8 hand configurations and 9 movements of the wrist. Then, we are going to solve a classification problem with 18 classes: 17 movements plus the rest posture.

\subsection{Features} \label{sec:Features}
A standard process is now applied to raw data~\cite{kuzborskij12, gijsberts_TNSRE_2014}.
The initial data consists of 12 signals, one for each electrode, composed of rest-movement sequences. We separate different postures and divide each one in overlapping sliding windows (10 ms) of fixed length (200 ms). Then, we separate training and test data.
The last step is feature extraction that, which determines the representation of the data for the subsequent classifier. We averaged the Mean Absolute Value, Variance and Waveform Length features based on their performance in prior studies~\cite{kuzborskij12} and to reduce the dependency on one specific type of representation.

\subsection{Classifiers} \label{sec:Classifier}
We build the classification models for the sources with a non-linear SVM with Gaussian kernel, using the LIBSVM library~\cite{CC01a}.
The adaptive algorithms used to build the classification models are described in Section \ref{sec:AdaptiveLearning}. We also considered two baselines: 
\begin{enumerate}
\item No Transfer: it uses an LS-SVM with Gaussian kernel trained on the target data only. This corresponds to learning to control the device without the help of prior knowledge.
\item Prior Features: it uses a linear LS-SVM trained on the prediction of source models only. This corresponds to attempting to control the prosthesis using only prior models, ignoring the target data.
\end{enumerate}
In the test phase, the accuracy for each model is given by the evaluation of the agreement between real and predicted labels of the test vectors.

All the classification models are based on a non-linear SVM with Gaussian kernel.
The SVM and Gaussian hyperparameters, respectively $C$ and $\gamma$, are set with cross validation on a grid with the following values $C \in \lbrace 0.01,0.1,1,10,100,1000 \rbrace$ and $\gamma \in \lbrace 0.01,0.1,1,10,100,1000 \rbrace$. 

\subsection{Protocol}
We ran three different experiments:

\begin{enumerate}
\item intact target subjects that exploit prior knowledge from other intact sources;
\item amputated target subjects that exploit prior knowledge from other amputated sources;
\item amputated target subjects that exploit prior knowledge from intact sources.
\end{enumerate}

To create each target's model we increase a set of randomly selected training vectors in steps of 120 samples, up to a maximum of 2160. We do not balance the vectors from different classes to test the quality of each model in the worst conditions. The performance is evaluated with about $2 \cdot 10^{4}$ test vectors per target. Thus, we consider the trend of accuracy averaged over subjects as a function of the number of training vectors.

\section{Experiments} \label{sec:Experiment}

\begin{figure*} [tbp]
\begin{center}
  {\centering \includegraphics [scale = 0.26] {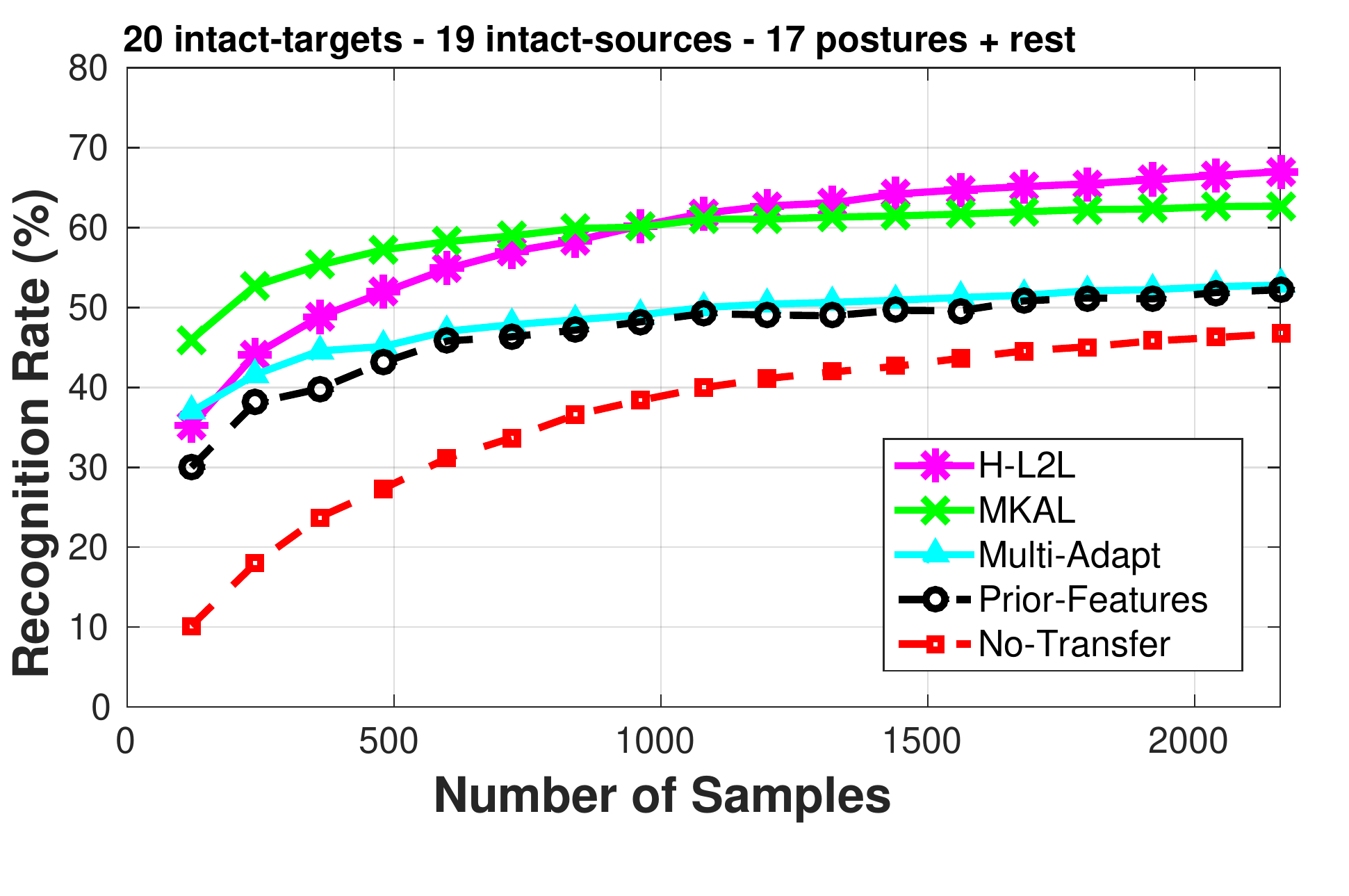}}\qquad
  {\centering \includegraphics [scale = 0.26] {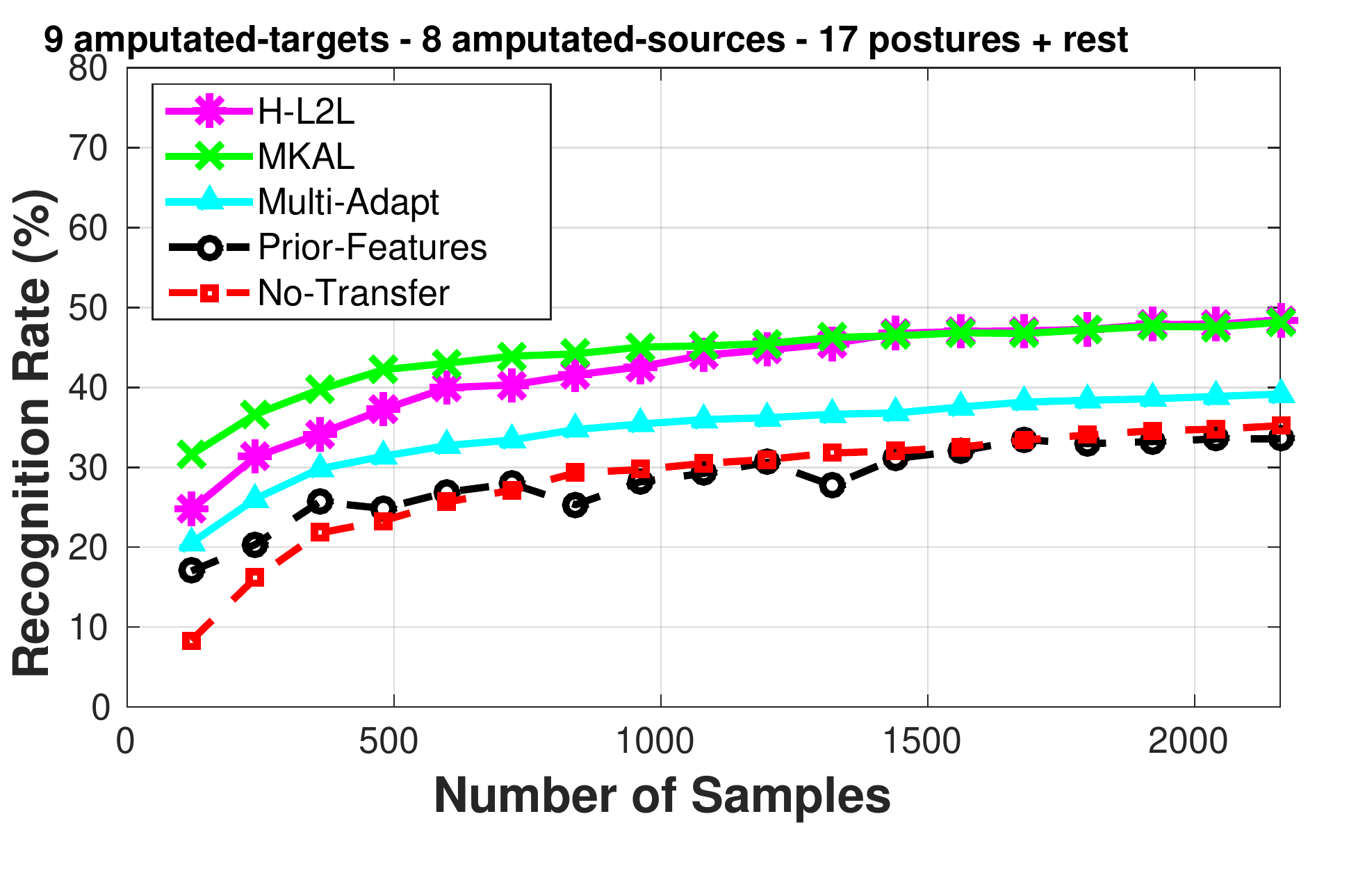}}\qquad
  {\centering \includegraphics [scale = 0.26] {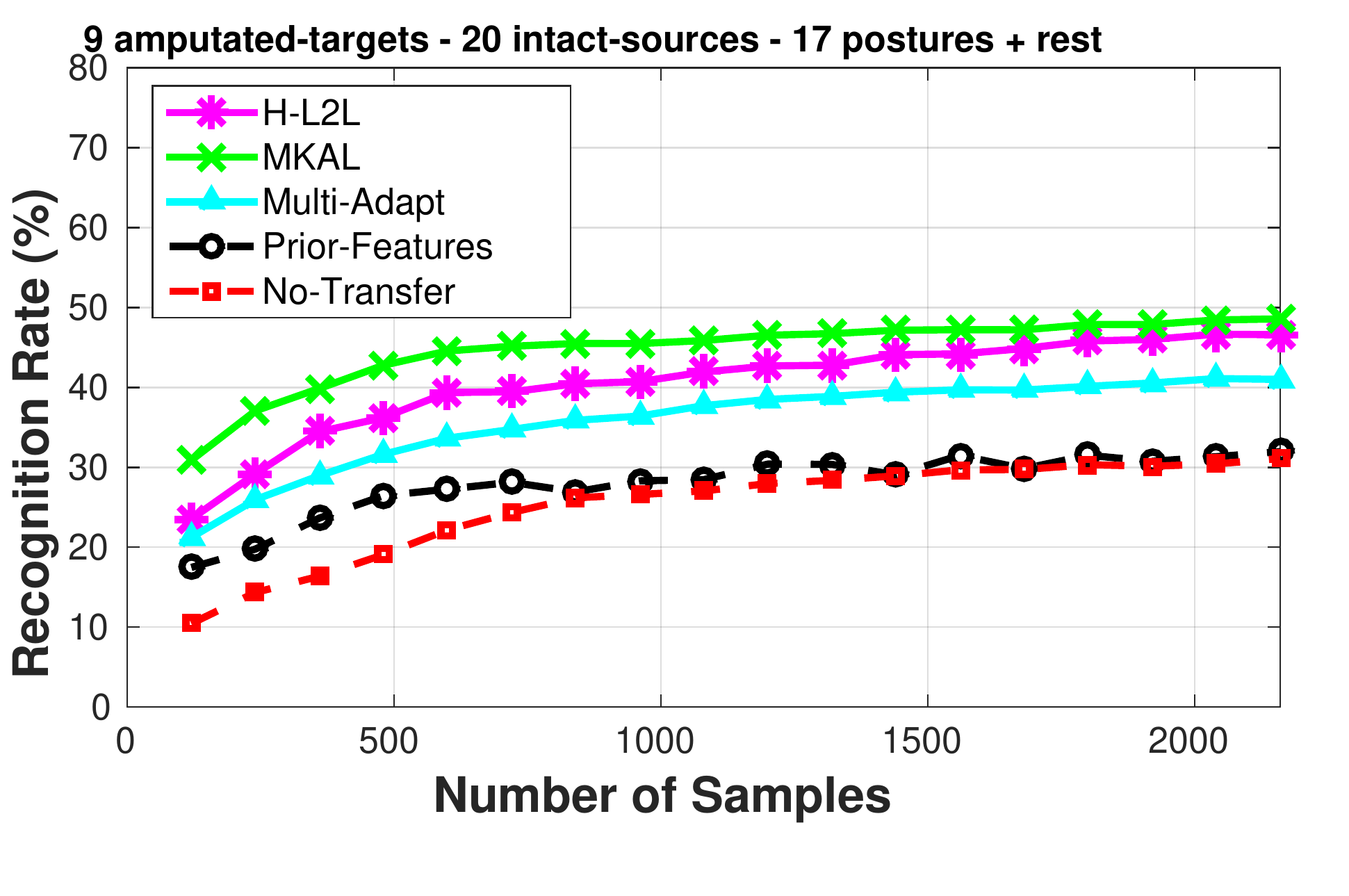}}\vspace{3pt}
  {\centering \includegraphics [scale = 0.26] {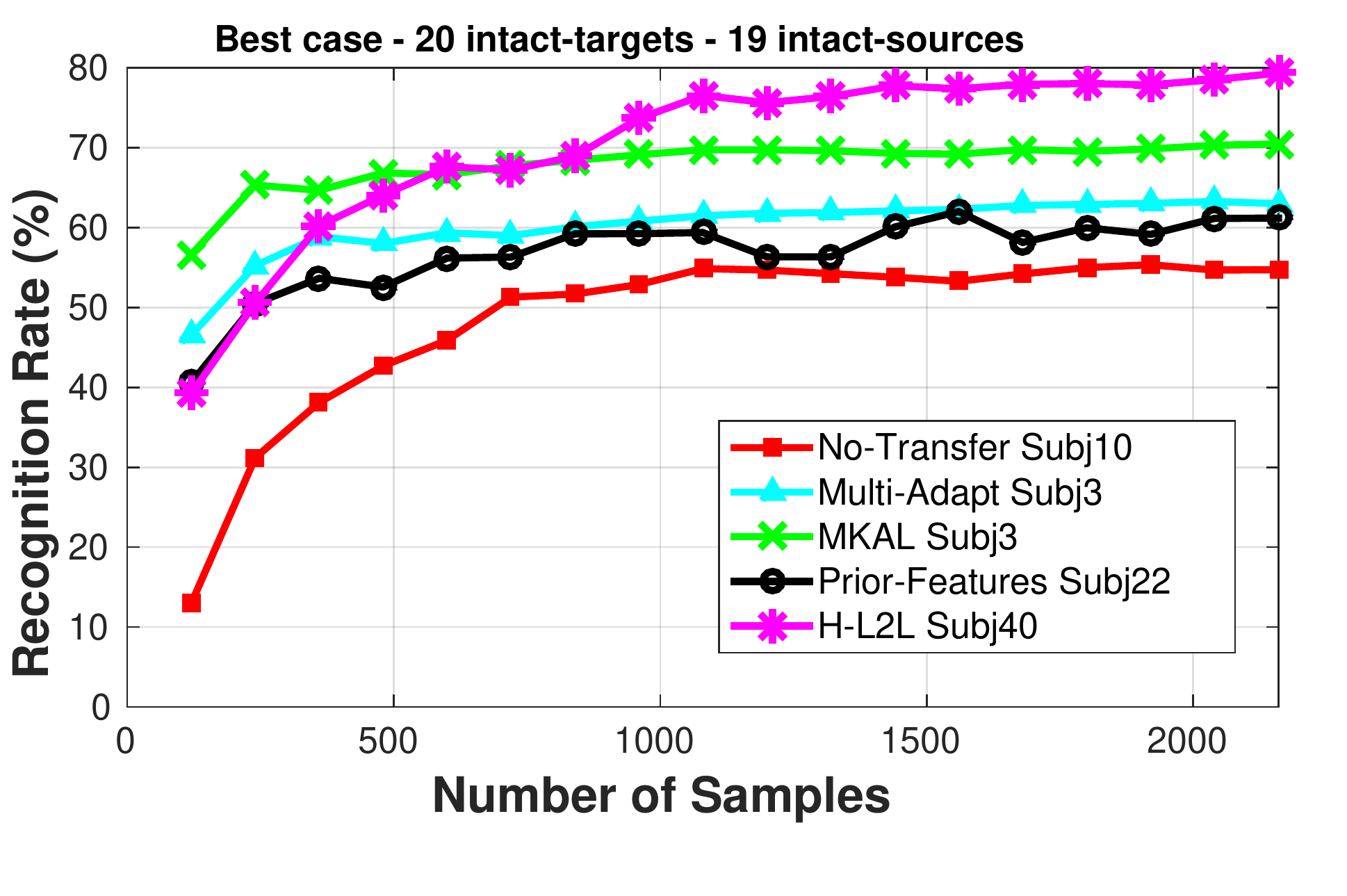}}\qquad
  {\centering \includegraphics [scale = 0.26] {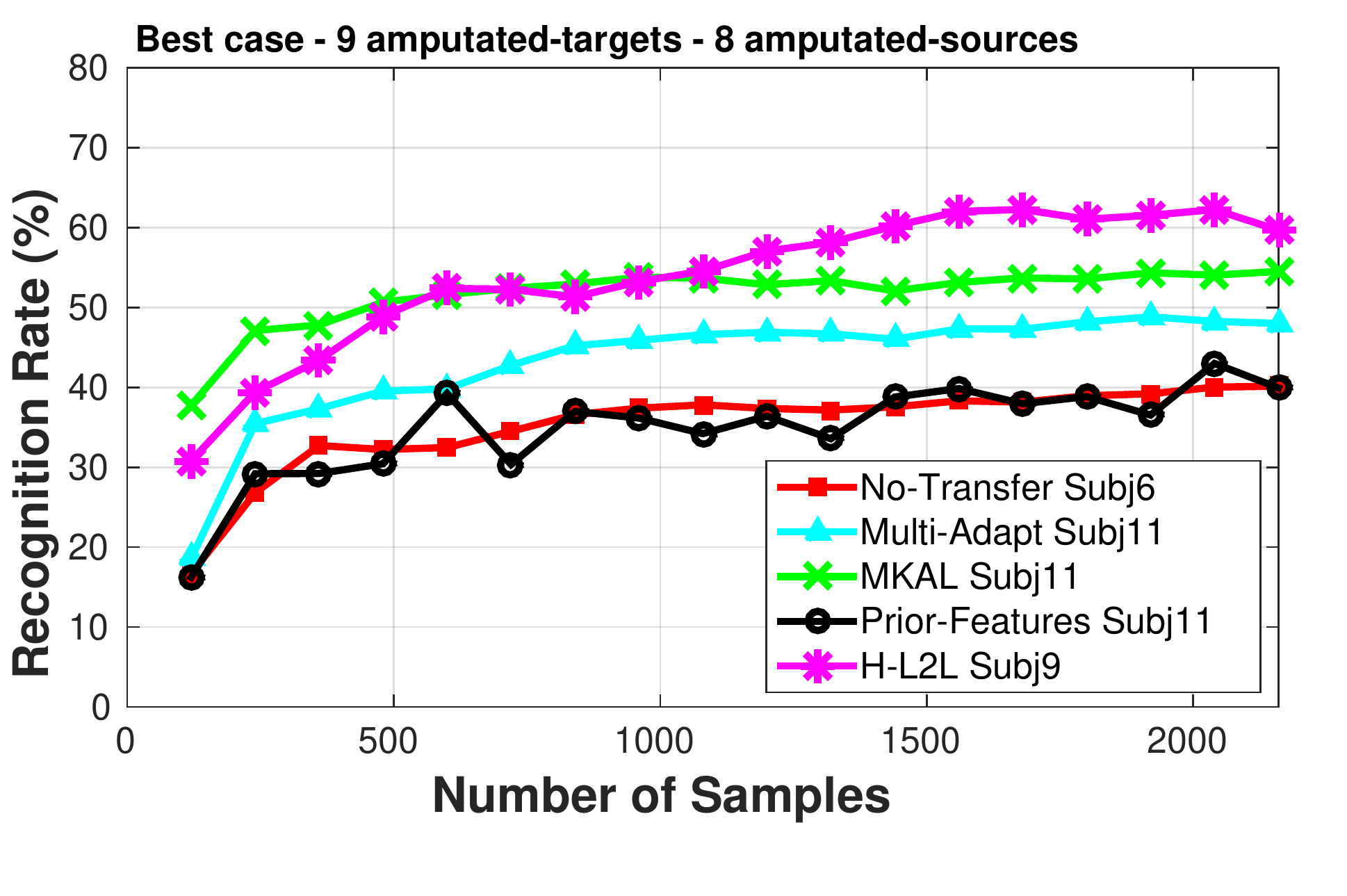}}\qquad
  {\centering \includegraphics [scale = 0.26] {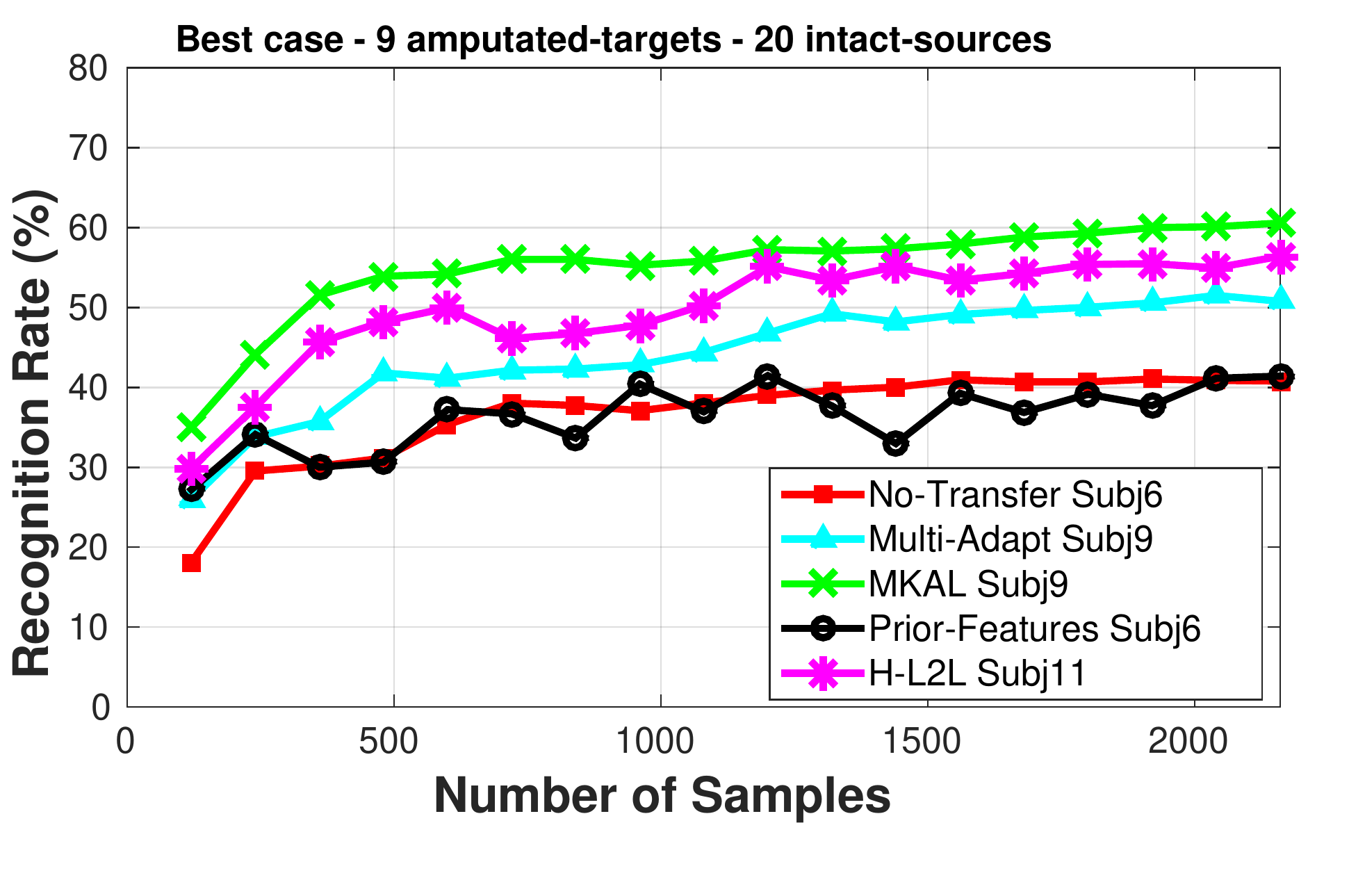}}\vspace{3pt}
  {\centering \includegraphics [scale = 0.26] {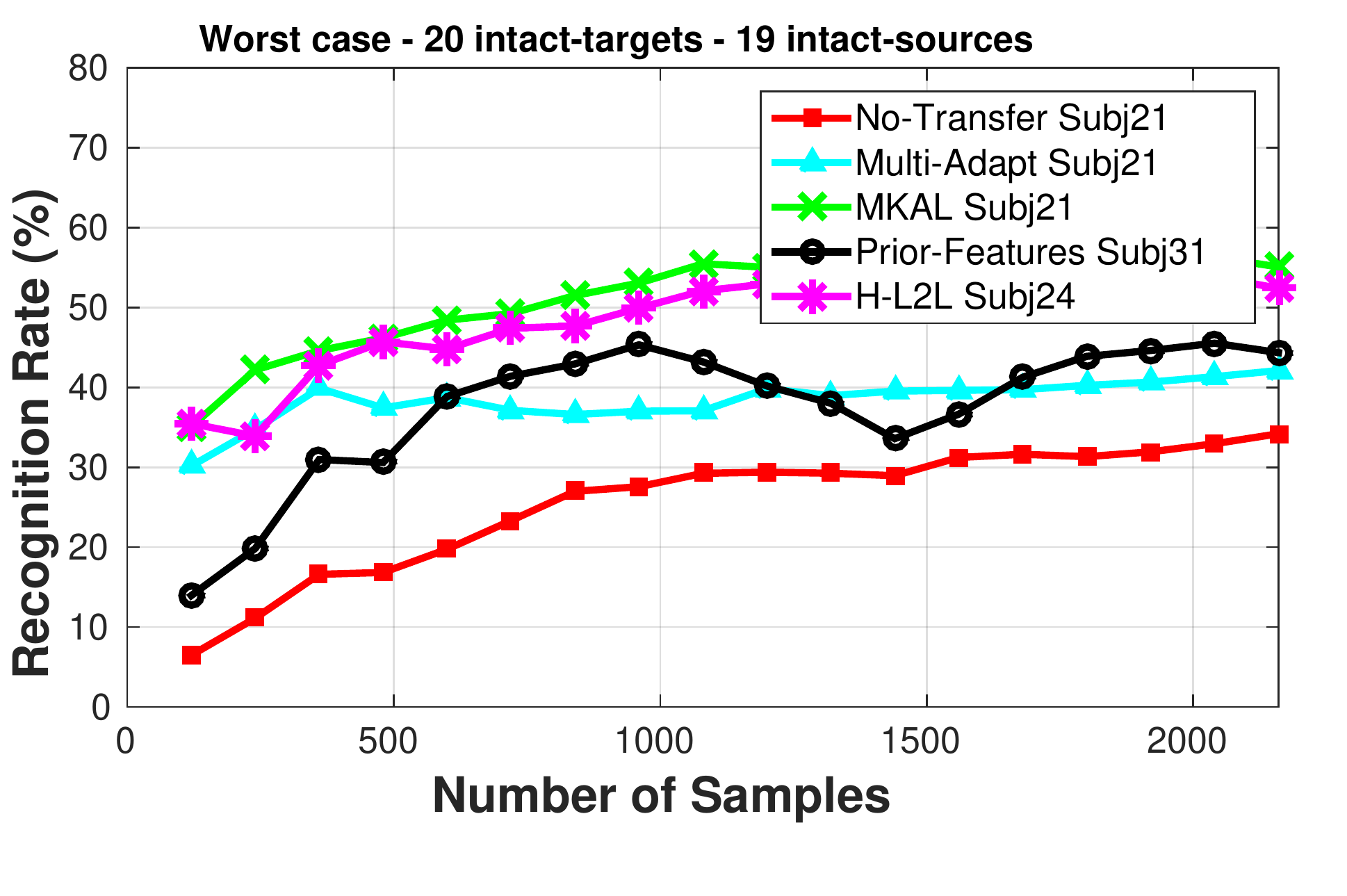}}\qquad
  {\centering \includegraphics [scale = 0.26] {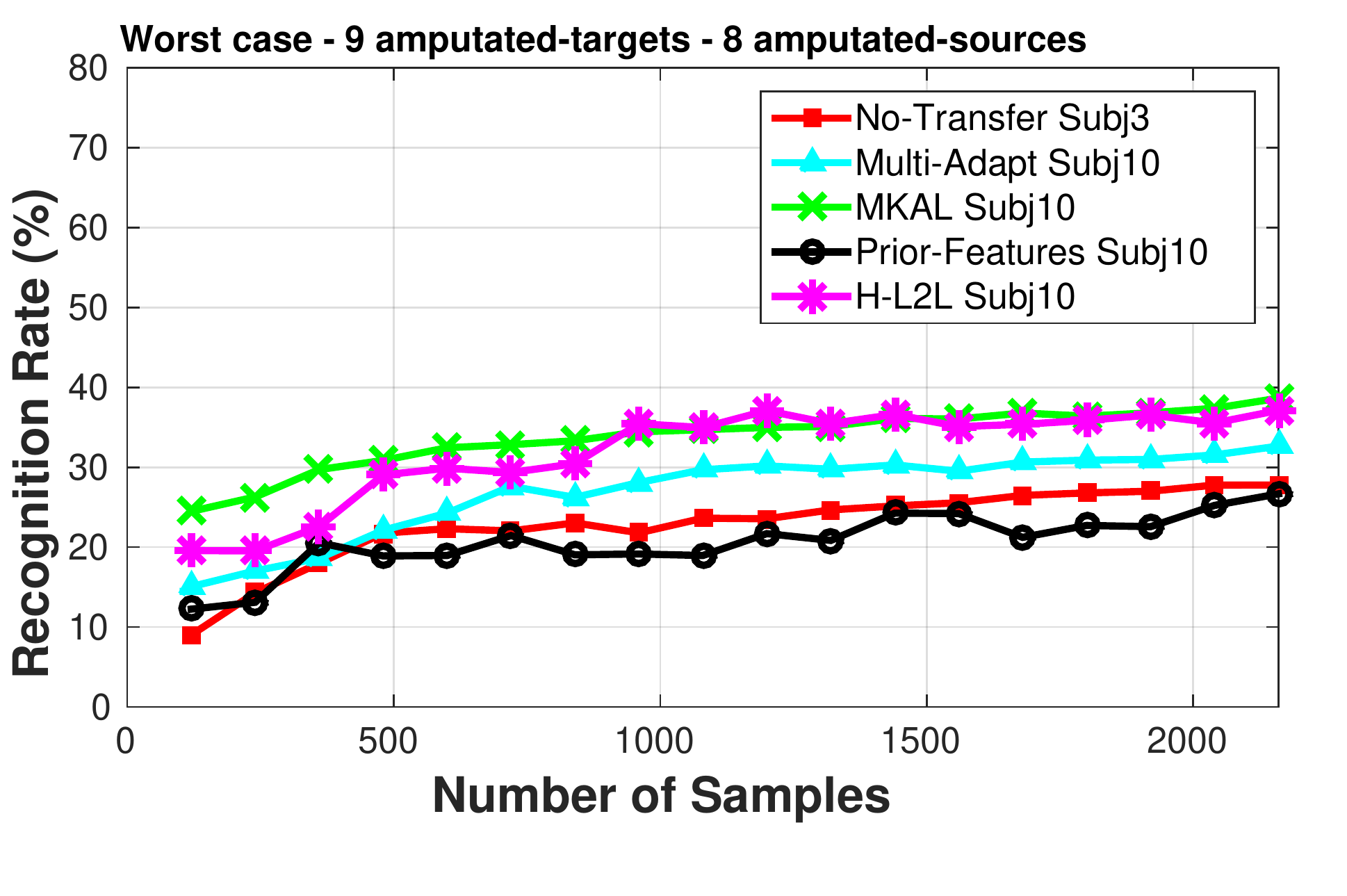}}\qquad
  {\centering \includegraphics [scale = 0.26] {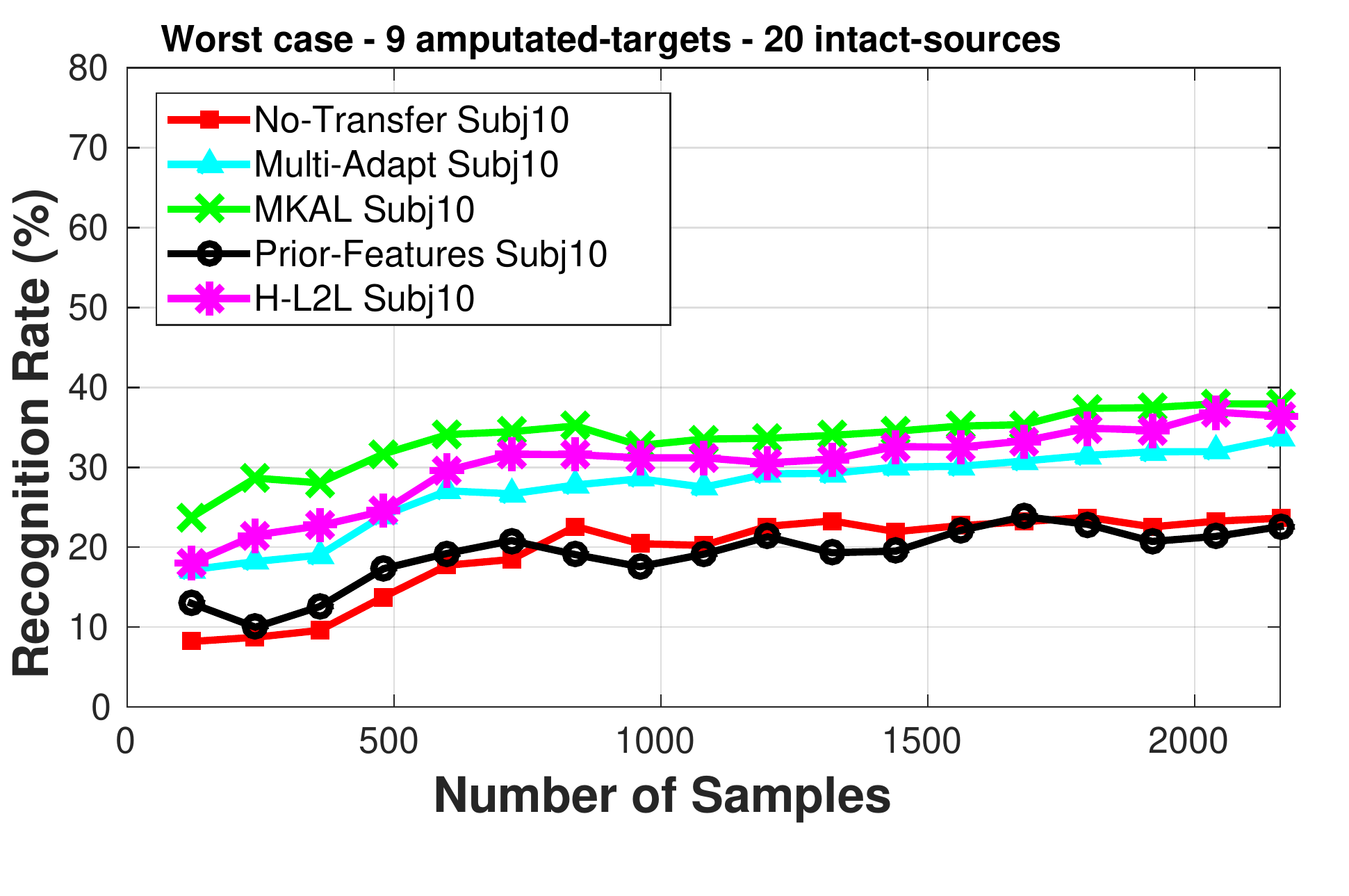}}\vspace{3pt}
\end{center}
  \caption{Classification rate obtained by averaging over all the subjects (top), for the best (middle) and for the worst cases (bottom) as a function of the number of samples in the training set: first experiment (left), second experiment (middle), third experiment (right).}\label{fig:RecRate}
\end{figure*}

We evaluate the results obtained using the algorithms described in Section \ref{sec:AdaptiveLearning} for data from the NinaPro database (Section \ref{sec:Data}) processed as explained in Section \ref{sec:Features}.
We ran three experiments. The first involves 20 random intact subjects, the second 9 no-random amputated subjects and the third both of them. In the first and second set-up, one by one, each subject represents the target problem just once, while the remaining are used as sources. In the last experiment the amputees are only the target, one by one, and the intact subjects represent only the sources.
The first experiment was performed also with 10 and 30 subjects, but taking into account that the final performance slightly increases from 10 to 20 and does not change from 20 to 30, in this work we report only the analysis for 20 subjects. 
In the second experiment we tested all the 11 amputated subjects, but 2 of them had only 10 electrodes because of insufficient space in the stump and they resulted lower in performance. Thus, in the following we report only the results for 9 subjects.

\subsection{Recognition Rate Analysis}
In Figure \ref{fig:RecRate} we report the average, minimum and maximum performance over the set of target subjects. Despite a component of noise in the best and worst cases, the ranking between different algorithms is preserved.
The domain adaptation algorithms always outperform No Transfer with an average of about: 12$\%$ (MA), 23$\%$ (MKAL) and 22$\%$ (H-L2L) in the first experiment; 6$\%$ (MA), 16$\%$ (MKAL) and 14$\%$ (H-L2L) in the second; 11$\%$ (MA), 20$\%$ (MKAL) and 15$\%$ (H-L2L) in the third. The knowledge of the sources positively affects the learning of a target, improving the performance even with few training samples. Furthermore, observe that the maximum level of performance of the No Transfer baseline is achieved by the domain adaptation algorithms using only a fraction of the target training data (i.e., from 240 up to 840 samples). In other words, prior knowledge allows us to reduce the training time by an order of magnitude, regardless of the nature of sources and target. 
The main difference between the three experiments concerns the absolute level of performance: the recognition rate of the first experiment is higher than the one of the second and third. This can partially be explained by the fact that classification is more difficult on EMG data from amputees (e.g., see \cite{atzoriScData14}), as demonstrated by the difference in performance of the No-Transfer baselines in the first two experiments.
Furthermore, observe that the relative improvement of domain adaptation algorithms is not as large for amputees as it is for intact subjects. This is a strong indicator that there is more variability in the EMG domains of amputees and it is therefore more difficult to find an overlap between the domain of an amputee and those of other subjects.

\subsection{Movement Classification Analysis}
\begin{figure*} [tbp]
  {\centering \includegraphics [scale = 0.27] {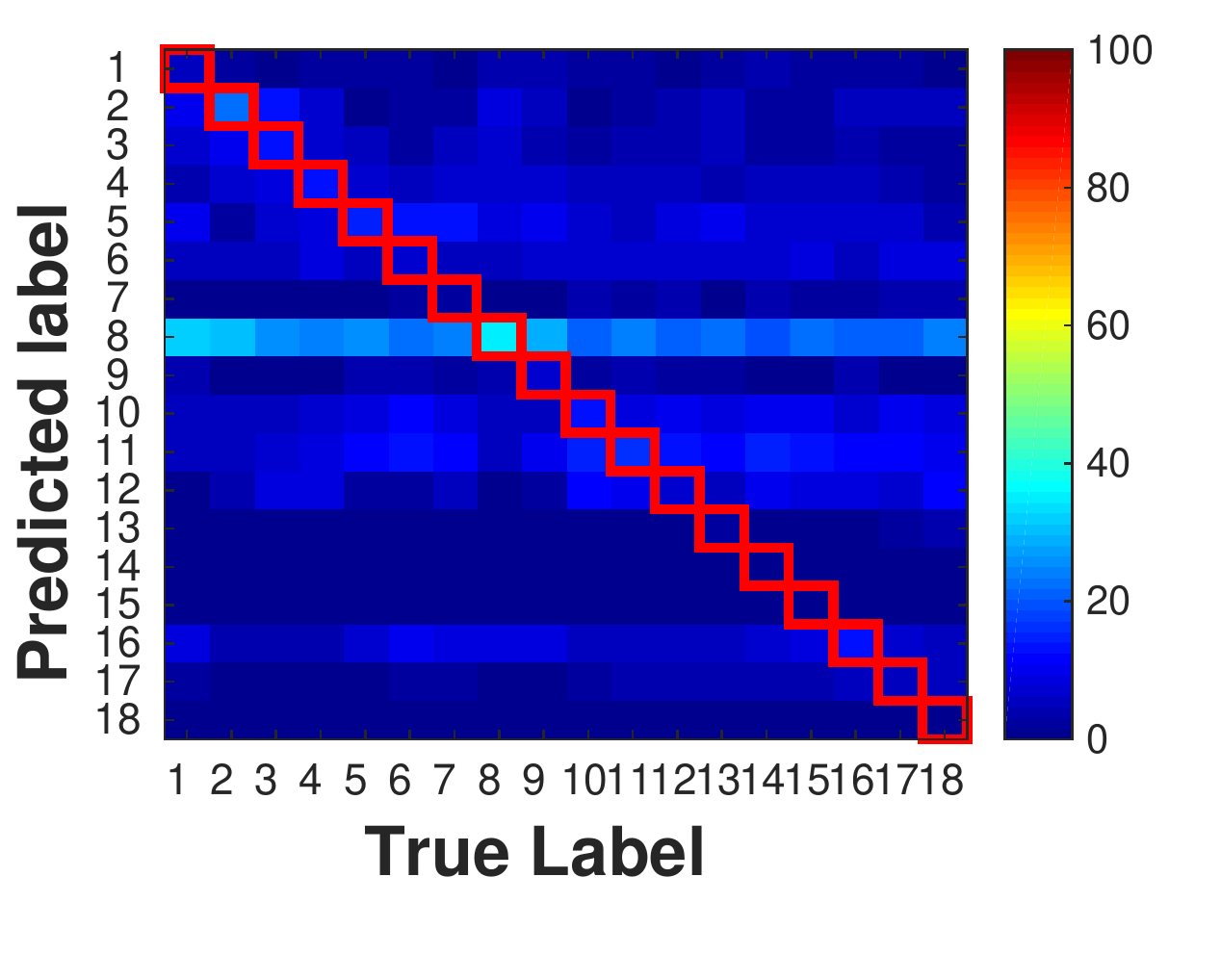}}\hspace{-2pt}
  {\centering \includegraphics [scale = 0.27] {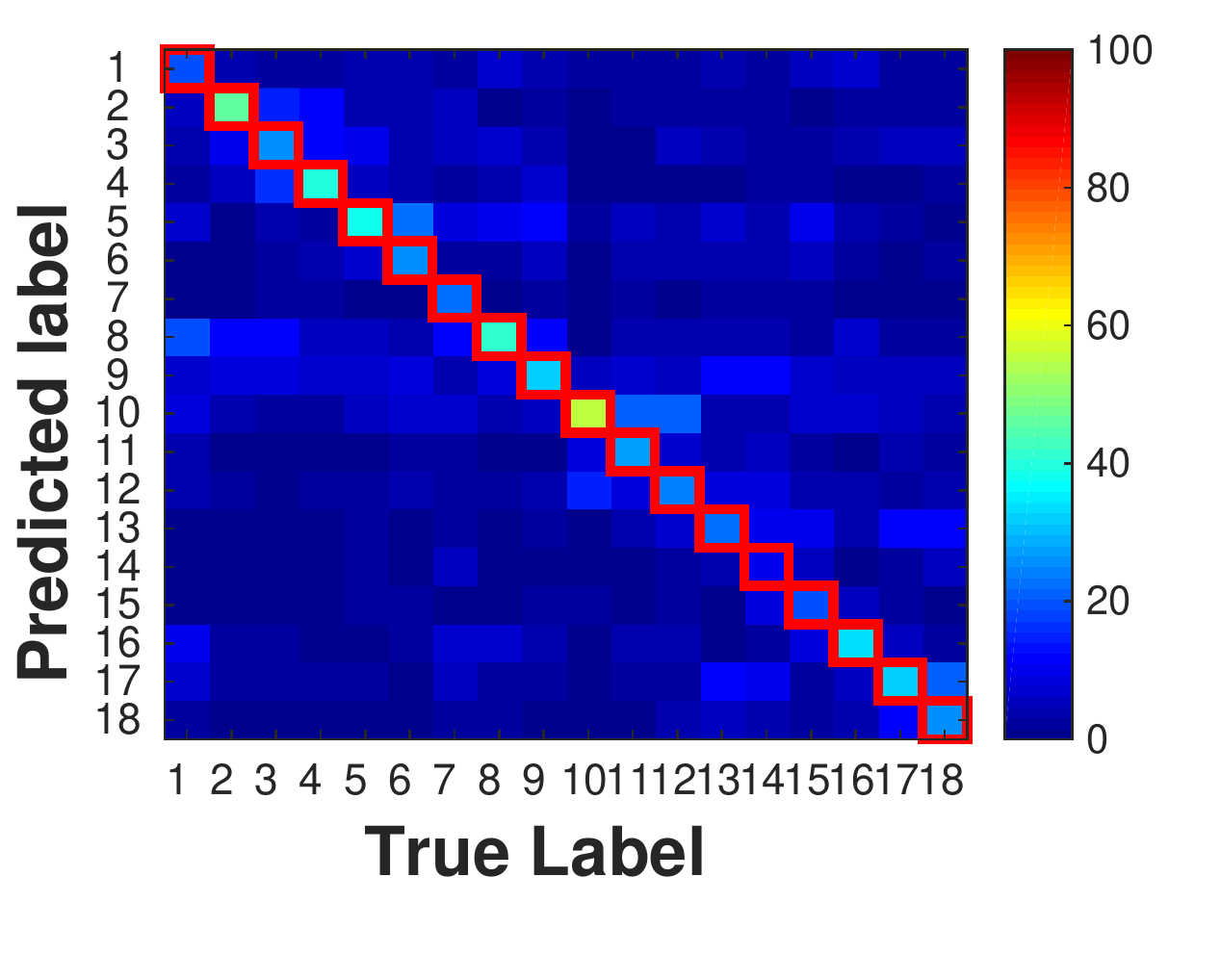}}\hspace{-2pt}
  {\centering \includegraphics [scale = 0.27] {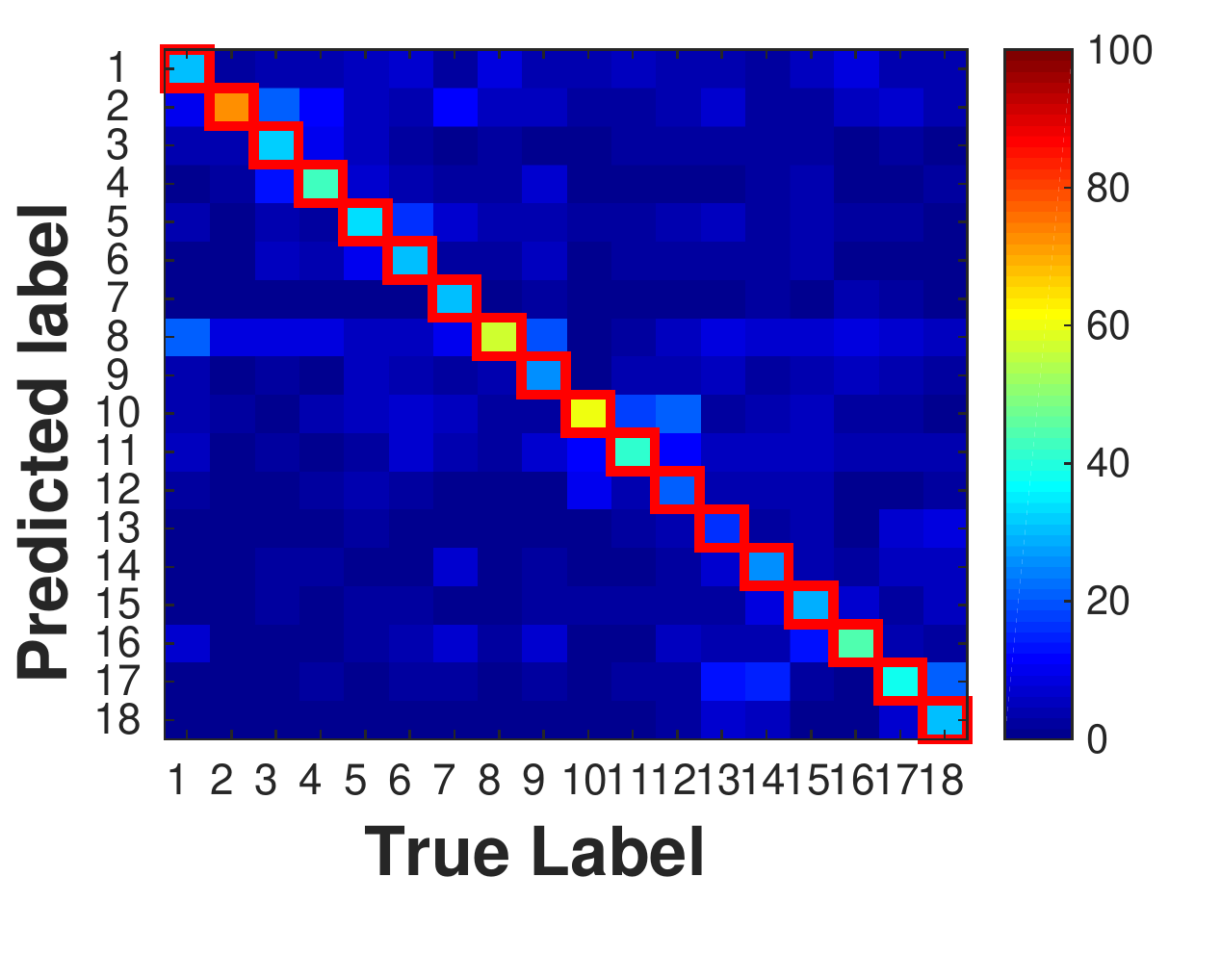}}\hspace{-2pt}
  {\centering \includegraphics [scale = 0.27] {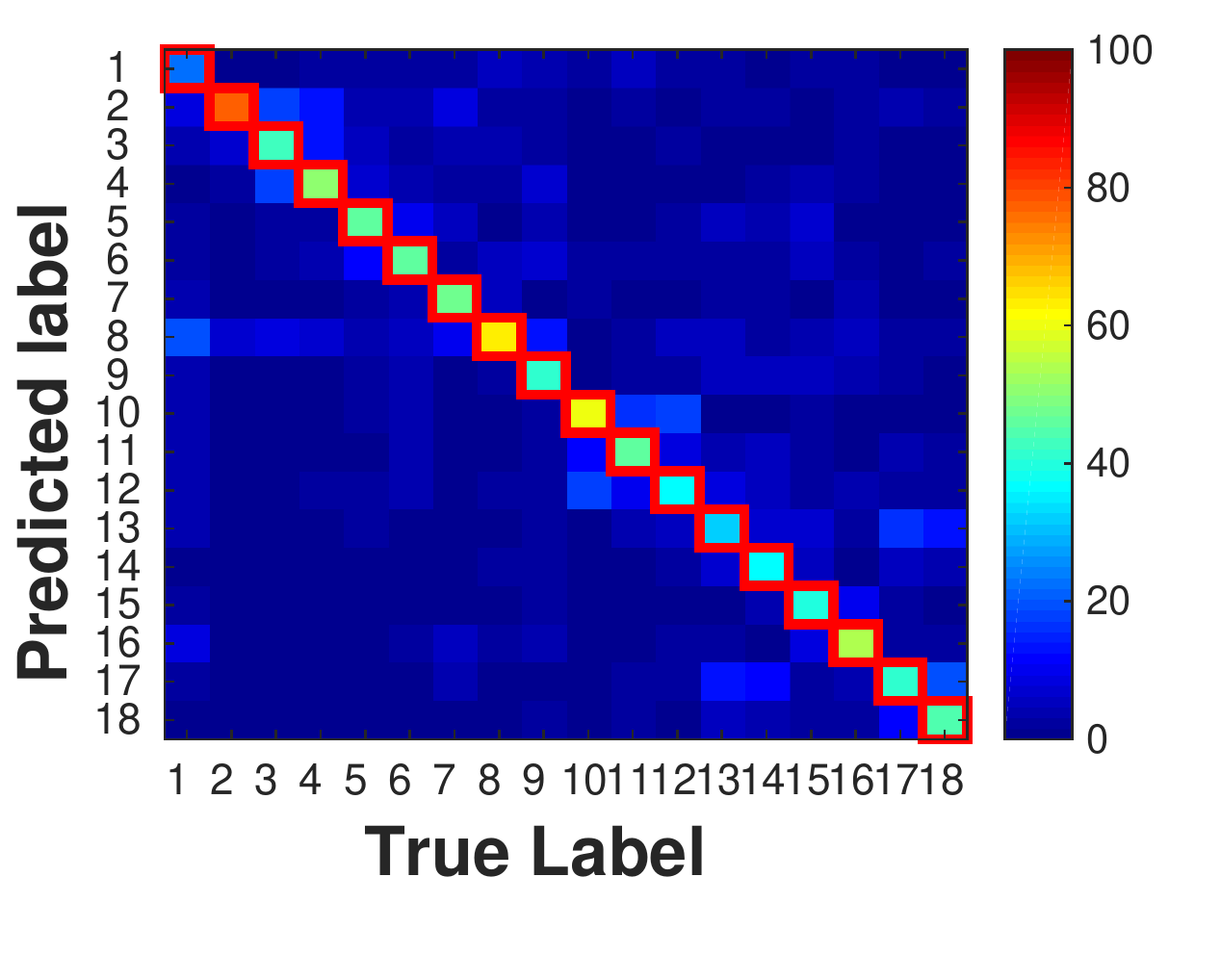}}\hspace{-2pt}
  {\centering \includegraphics [scale = 0.27] {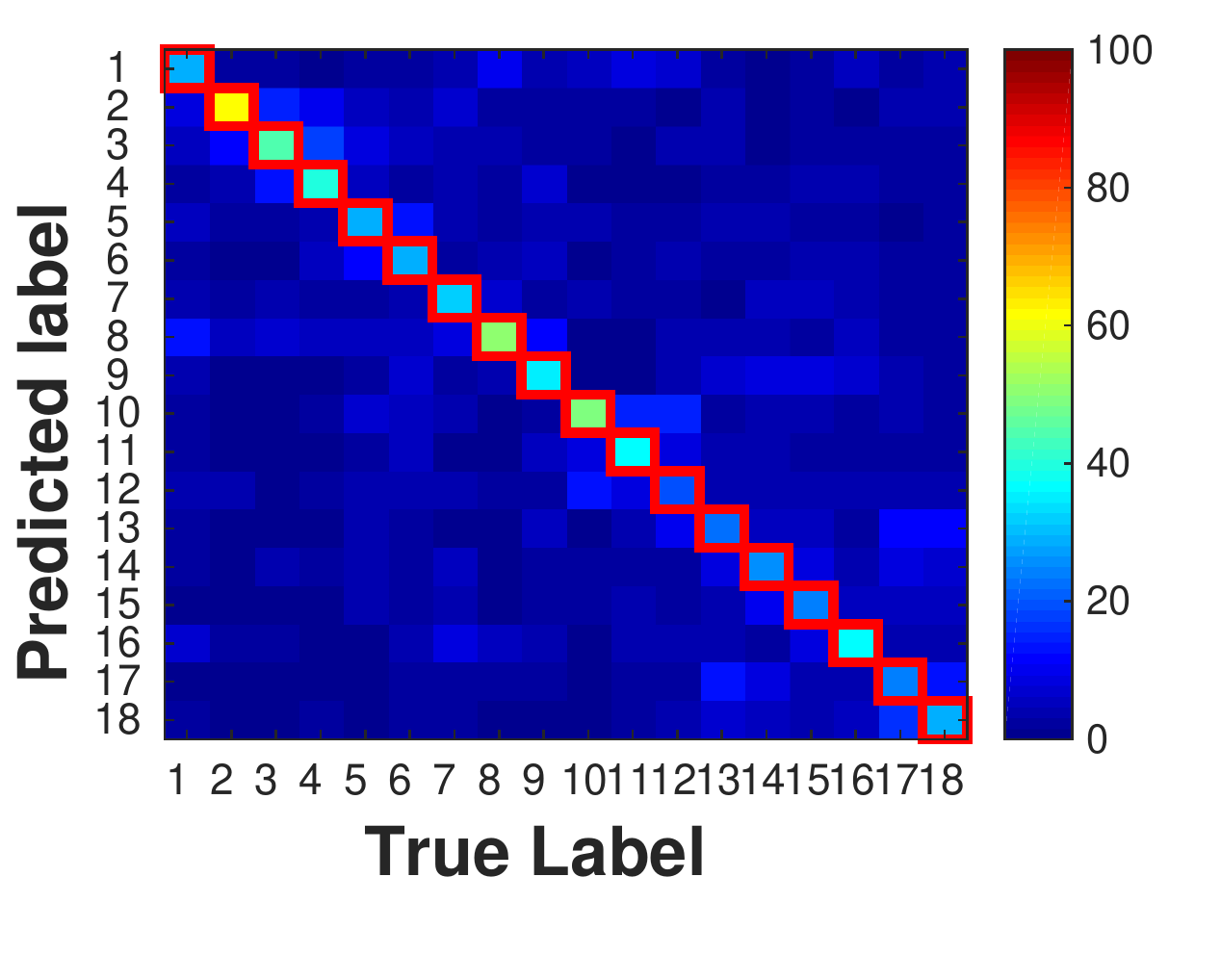}}
  \caption{Confusion matrices for (from left to right) No Transfer, Prior Features, MA, MKAL and H-L2L from the first experiment with 120 training samples.}\label{fig:ConfMat}
\end{figure*}

The recognition rate gives information about the average classification accuracy, ignoring the exact type of misclassifications by an algorithm. To investigate this recognition behaviour of each class in more depth, we  show the confusion matrices in Figure \ref{fig:ConfMat}. In the ideal case the matrix is red in the diagonal and blue outside: the predicted labels are equal to the true ones and the wrong predictions are equal to zero. 
We report the results with only 120 training samples to highlight the improvement when exploiting prior knowledge in the small sample setting.
Focusing on No Transfer we can observe its strong dependence on the specific set of random training vectors. With only few samples we obtain a classifier that is biased toward the movements that appeared most frequently in the training data. Prior knowledge increases the recognition for all classes: even with few training samples the highest mean prediction corresponds to the true labels for the majority of movements.

\begin{figure*} [tbp]
  {\centering \includegraphics[scale=0.26]{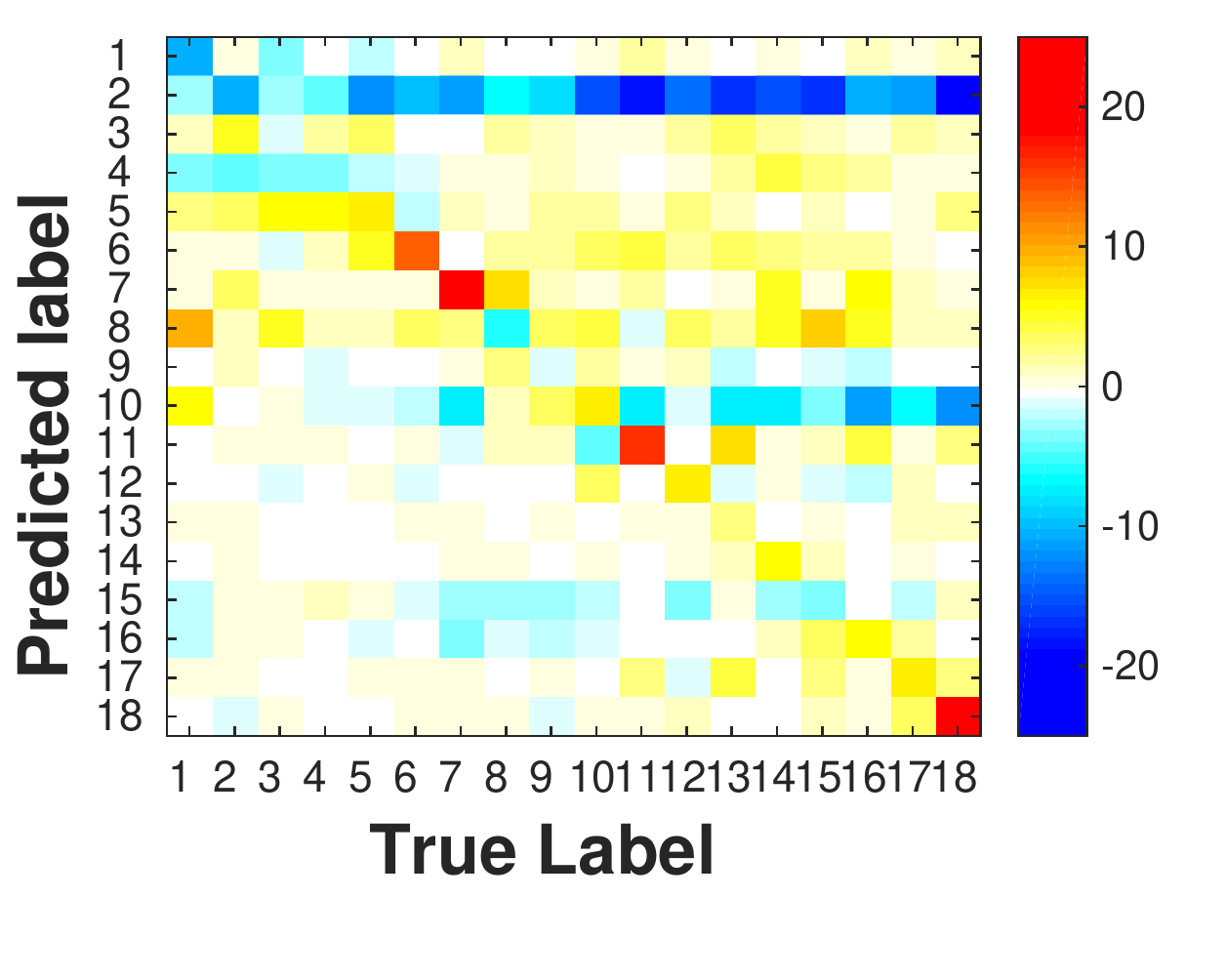}} \hspace{-1pt}
  {\centering \includegraphics[scale=0.26]{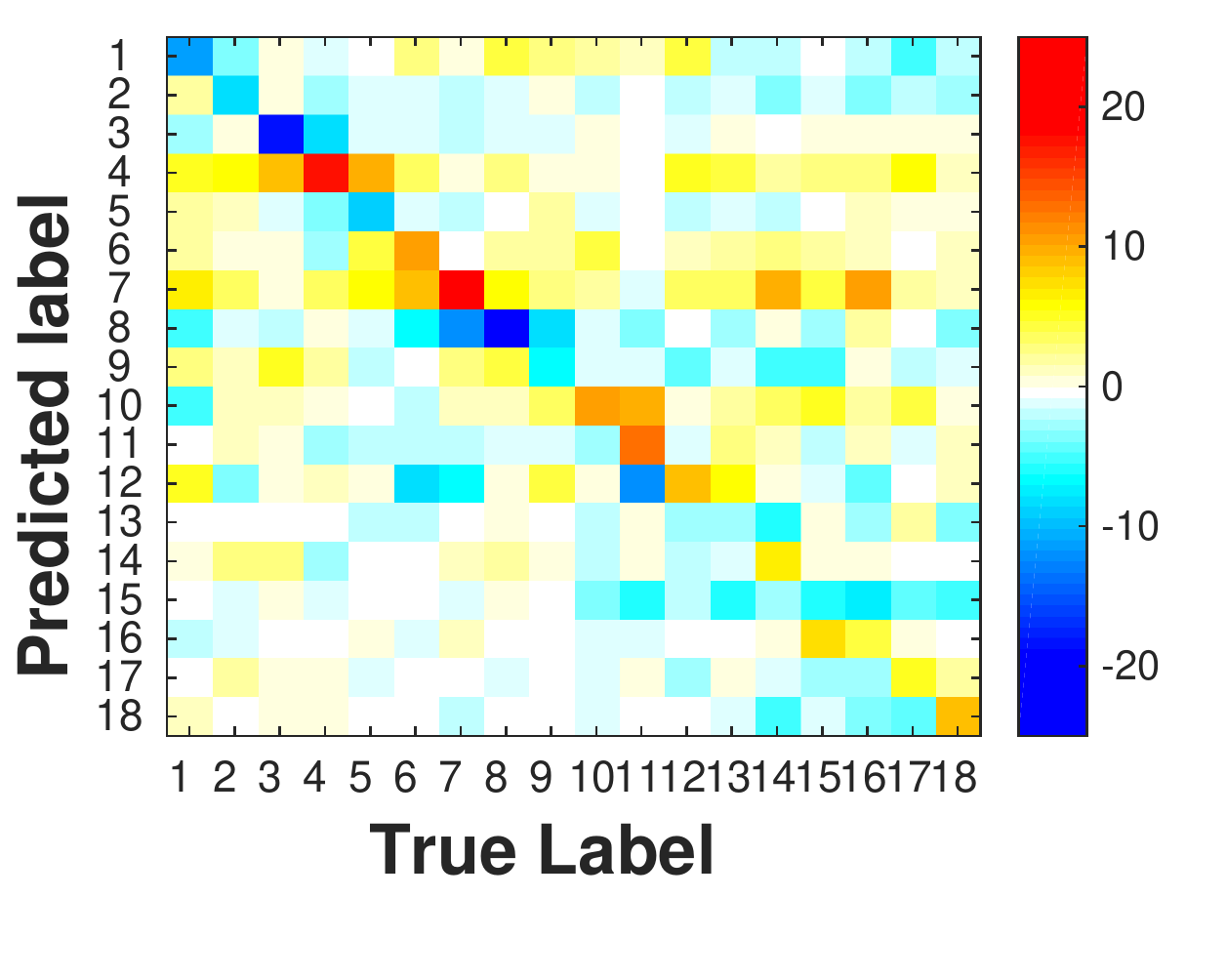}} \hspace{-1pt}
  {\centering \includegraphics[scale=0.26]{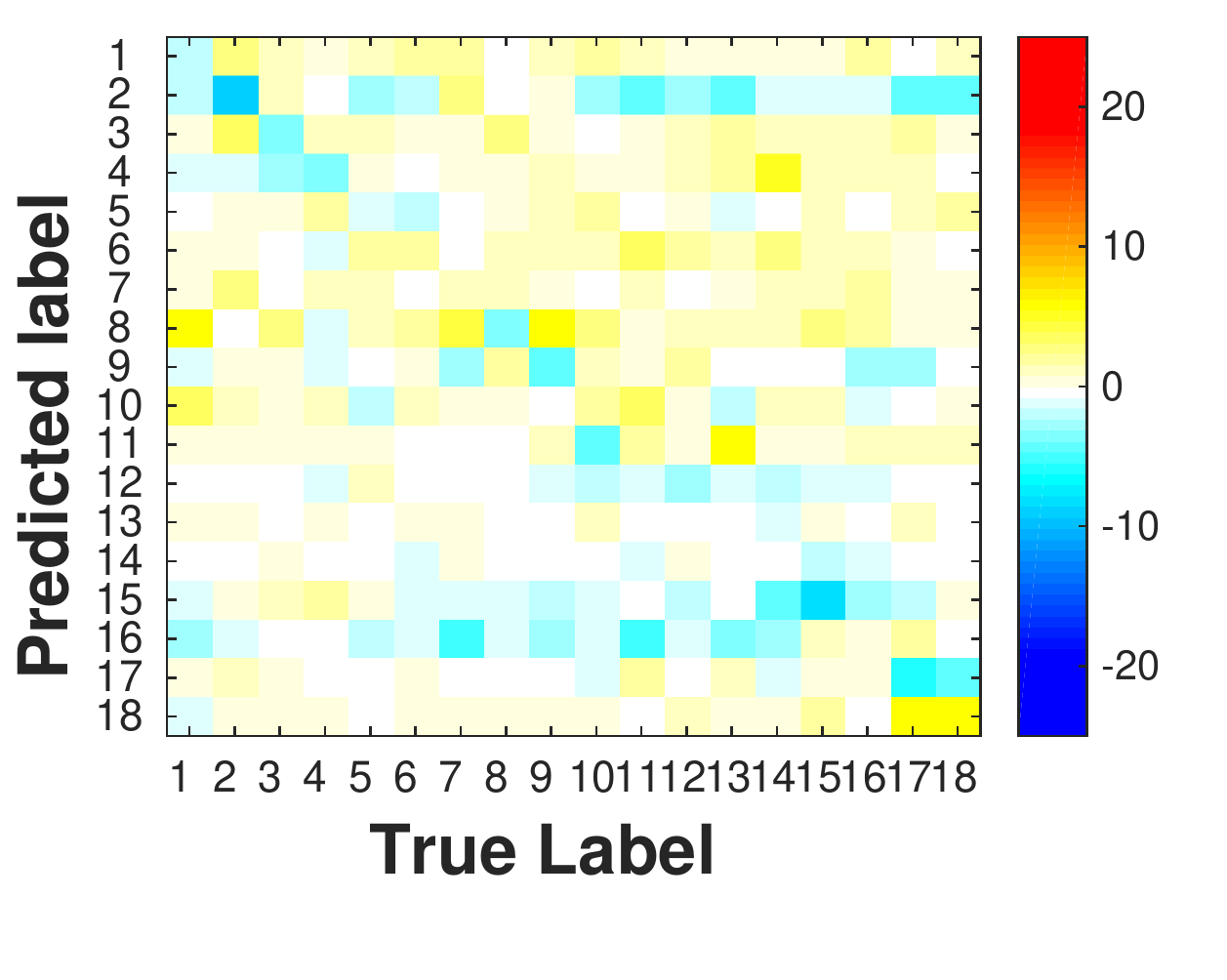}} \hspace{-1pt}
  {\centering \includegraphics[scale=0.26]{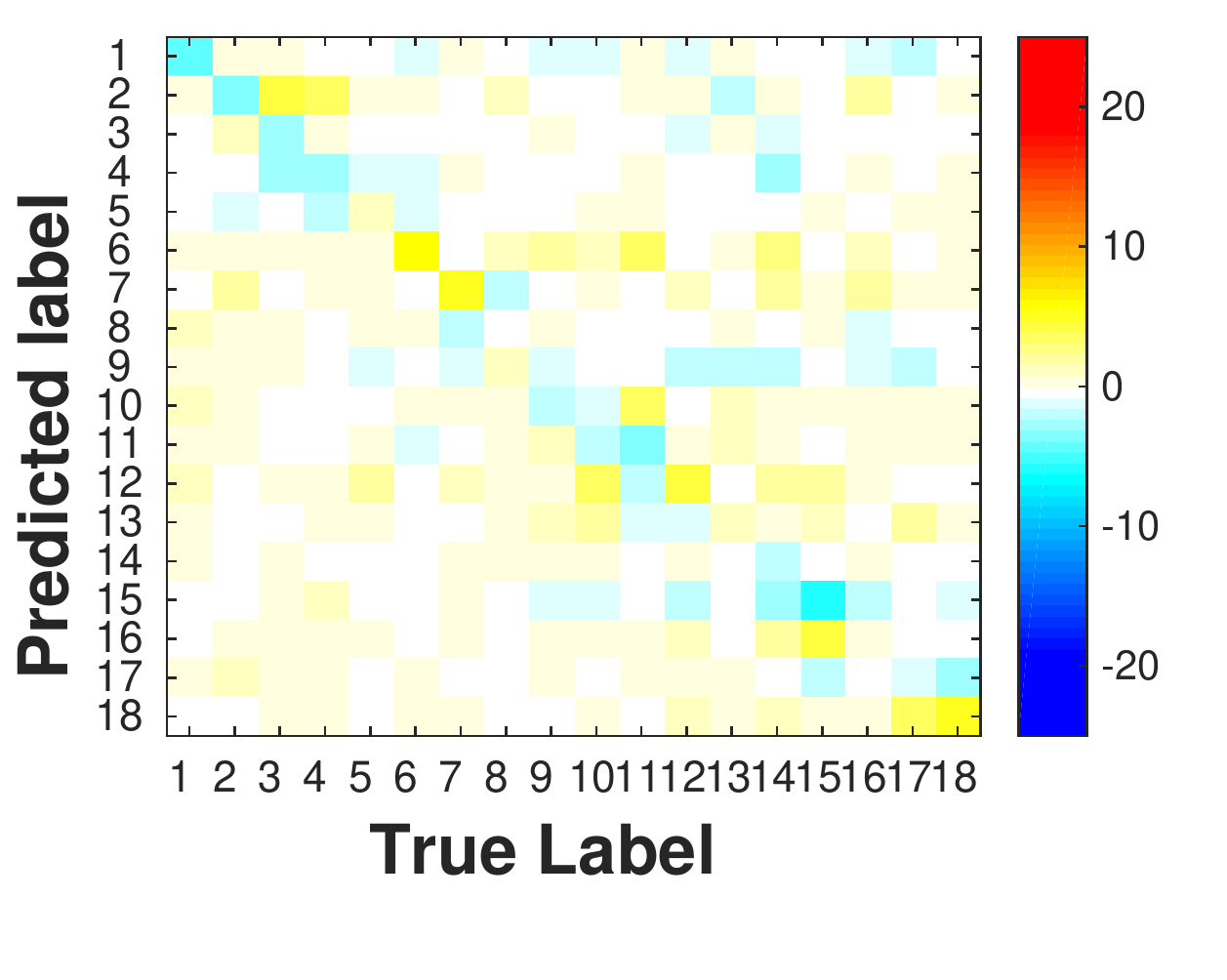}} \hspace{-1pt}
  {\centering \includegraphics[scale=0.26]{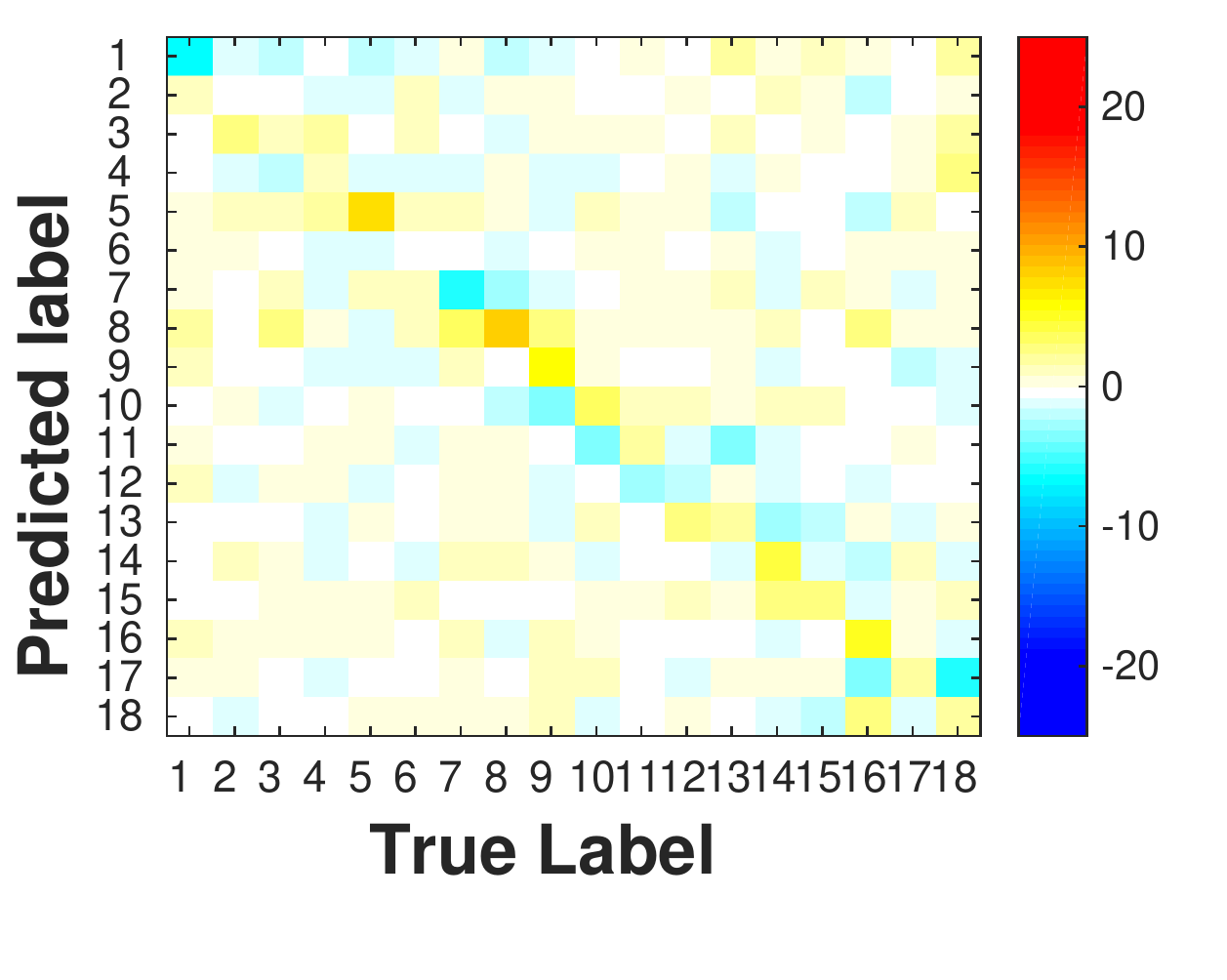}}
  \caption{Difference between the confusion matrices of the second and third experiment for (from left to right) No transfer, Prior, MA, MKAL and H-L2L with 2160 training samples.}	\label{fig:DiffConfMat}
\end{figure*}

It is also interesting to focus on the difference of confusion matrices of the second and third experiment for the same algorithm: red, blue and white colours represent respectively an increase, decrease and absence in (mis)classification (Figure \ref{fig:DiffConfMat}).
Looking at the diagonal there are some movements that appear differently recognized in the two experiment, but this difference varies in a range of [-9$\%$,6$\%$] for MA, [-6$\%$,6$\%$] for MKAL and [-7$\%$,8$\%$] for H-L2L exploiting all the training samples available. This highlights that an amputee does not suffer a significant learning gap exploiting intact or amputated sources. Note that the discrepancies in the No-Transfer case are caused by a different instantiation of the set of training samples. In fact, we also observe that these differences are much smaller in case of the domain adaptation algorithms, showing that using prior information has a stabilizing effect on the learning.

Focusing on single movement recognition, we want to understand the similarity in misclassifications when varying the domain adaptation algorithm or the number of training samples. For this purpose we take into account the four classes with the highest predictions for each real class (i.e., each column of the confusion matrices). We ignore the fifth prediction as it usually has a recognition lower than 5$\%$. 
Comparing different algorithms for the same number of training samples, we calculate the percentage of real classes that are equal in at least 3 out of the 4 most common (mis)predictions (Table \ref{tab:PercTab1}). We did the same analysis comparing the predictions for the same algorithm with a different number of training vectors (Table \ref{tab:PercTab2}).
The findings show that a class is generally misclassified with the same set of postures, regardless of the number of training samples and the algorithm used.

\begin{table}[!t]
	\centering
	\resizebox{8.5cm}{!}	{
	
	\begin{tabular} {|| c | c | c | c | c ||}
	
	\hline
	\hline
	 Experiment & Training Vectors  & MA - MKAL & MKAL - H-L2L & MA - H-L2L \\ 		\hline
	\hline
	First & 120 & 72$\%$ ($13/18$) & 78$\%$ ($14/18$) & 83$\%$ ($15/18$) \\   \hline
	First & 1080 & 94$\%$ ($17/18$) & 100$\%$ ($18/18$) & 78$\%$ ($14/18$) \\   \hline
    First & 2160 & 78$\%$  ($14/18$) & 83$\%$ ($16/18$) & 72$\%$ ($13/18$) \\   \hline\hline
    Second & 120 & 61$\%$  ($11/18$) & 67$\%$ ($12/18$) & 33$\%$ ($6/18$) \\   \hline
	Second & 1080 & 72$\%$  ($13/18$) & 94$\%$ ($17/18$) & 72$\%$  ($13/18$) \\   \hline
    Second & 2160 & 72$\%$  ($13/18$) & 94$\%$ ($17/18$) & 56$\%$ ($10/18$) \\   \hline\hline
    Third & 120 & 50$\%$  ($9/18$) & 72$\%$ ($13/18$) & 33$\%$ ($6/18$) \\   \hline
	Third & 1080 & 89$\%$  ($16/18$) & 83$\%$ ($15/18$) & 67$\%$ ($12/18$) \\   \hline
    Third & 2160 & 72$\%$  ($13/18$) & 89$\%$ ($16/18$) & 61$\%$ ($11/18$) \\   \hline
	\end{tabular}
	
	}
	\vspace{3px}
	\caption{Percentage of similarity in the prediction of classes between the different adaptive algorithms for the same number of training vectors.}
	
\label{tab:PercTab1}
\end{table}

\begin{table}[!t]
	\centering
	\resizebox{8.5cm}{!}	{
	
	\begin{tabular} {|| c | c | c | c | c ||}
	
	\hline
	\hline
	 Experiment & Training Vectors  & MA & MKAL & H-L2L \\ 		\hline
	\hline
	First & 120-1080 & 94$\%$   ($17/18$) & 89$\%$ ($16/18$) & 78$\%$ ($14/18$) \\    \hline
	First & 1080-2160 & 100$\%$ ($18/18$) & 94$\%$ ($17/18$) & 94$\%$ ($17/18$)\\   \hline\hline
    Second & 120-1080 & 33$\%$  ($6/18$) & 72$\%$  ($13/18$) & 61$\%$  ($11/18$) \\    \hline
	Second & 1080-2160 & 94$\%$ ($17/18$) & 100$\%$ ($18/18$) & 94$\%$ ($17/18$)\\   \hline\hline
	Third & 120-1080 & 39$\%$ ($7/18$) & 83$\%$ ($15/18$) & 72$\%$ ($13/18$) \\    \hline
	Third & 1080-2160 & 78$\%$ ($14/18$) & 100$\%$ ($18/18$) & 89$\%$ ($16/18$)\\   \hline
	
	\end{tabular}
	
	}
	\vspace{3px}
	\caption{Percentage of similarity in the prediction of classes when the number of training vectors changes for the same methods.}
	
\label{tab:PercTab2}
\end{table}

\begin{figure} [tbp]
  \centering \includegraphics [scale = 0.25] {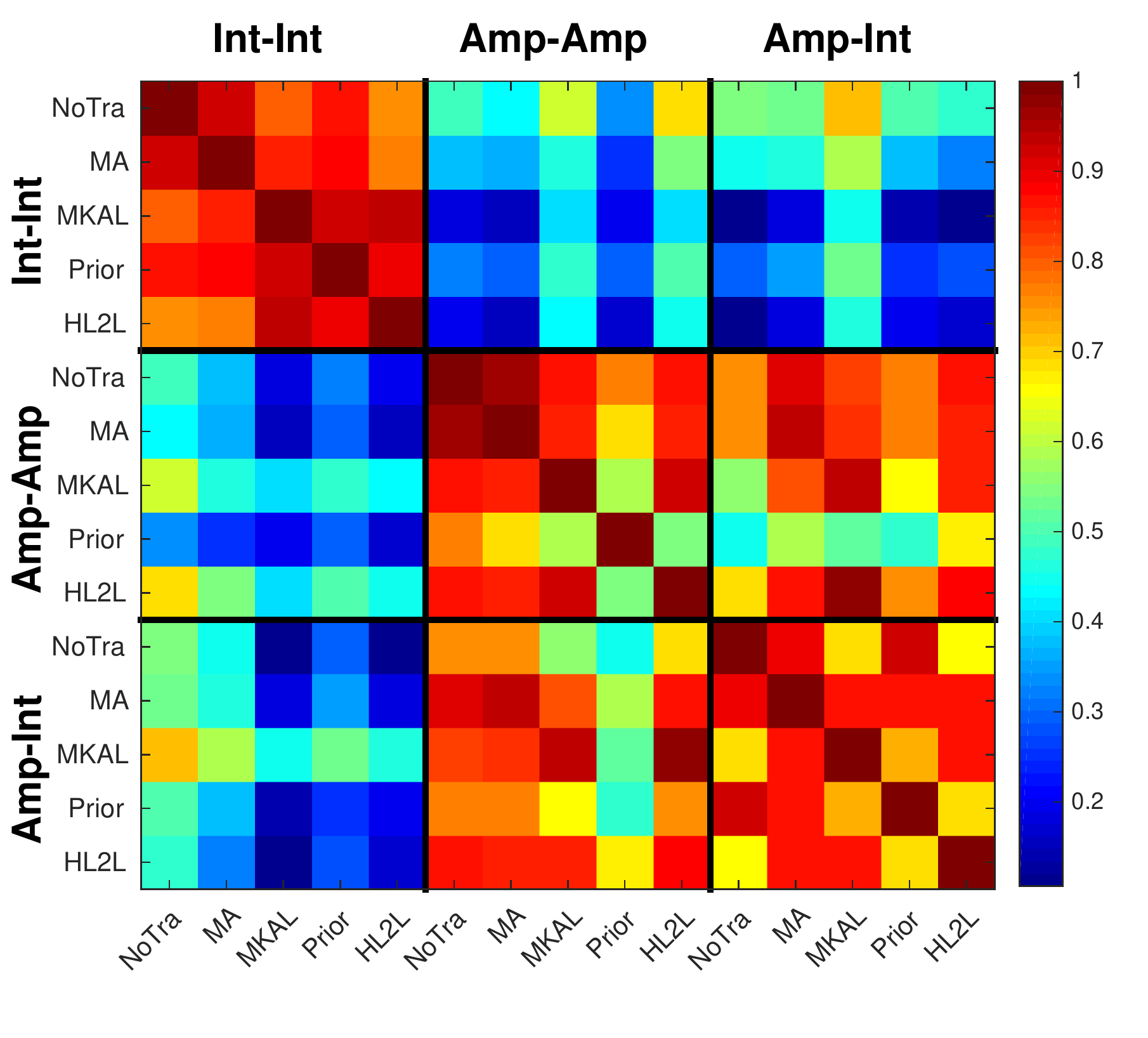} 
  \caption{Correlation matrix for each algorithm and experimental setting. II, AA and AI stand respectively for: intact-intact (i.e., first experiment), amputees-amputees (i.e., second experiment) and amputees-intact (i.e., third experiment).}\label{fig:CorrMat}
\end{figure}

Finally, we want to quantify the similarity between the confusion matrices of all the experiments and classification methods to understand if a posture is learned differently when changing the target and the sources. 
To this end we calculate the correlations between the average recognition percentage for each class when keeping the number of training samples fixed. This correlation is computed for each combination of experiment and method, resulting in the correlation matrix shown in Figure \ref{fig:CorrMat}.
The two diagonal blocks of the matrix contain the correlations between intact targets (first experiment) and amputated targets (second and third experiments). The blocks outside the diagonal hold the correlation between intact and amputated targets.
These findings show that the correlation between the second and third experiment is high, thus an amputee learns a movement with the same simplicity or difficulty from intact and amputated sources. Instead the correlation between amputated and intact targets is low, meaning that the set of movements that can be learned with relative ease is not identical in the two cases.

\section{Conclusion and Comparison} \label{sec:Conclusion}
In this work we have dealt with the control of non-invasive myoelectric prostheses.
We have investigated the variation in learning performance when prior information from other subjects is used and whether the type of knowledge (from amputees or intact) is significant for an amputee.
Our findings show that prior knowledge positively affects the trend of the recognition curves, increasing the performance and reducing the training time by an order of magnitude.
The main result concerns the comparison when using either intact or other amputated subjects as sources for an amputated target. The obtained mean performance appears unchanged for all the algorithms, meaning that an amputee learns with the same effectiveness from intact or amputated sources. The same conclusion emerges from the correlation matrices that show a high agreement between both settings: an amputee learns a movement with the same simplicity or difficulty from intact and amputated subjects. 
This result is important since there is a large number of potential intact source subjects with respect to amputees, for who participating in data collection is difficult of even painful.

\FloatBarrier
{\footnotesize
\bibliographystyle{IEEEtranN} 
\bibliography{bibICRR2017}
}

\end{document}